\documentclass[a4paper,fleqn]{cas-dc}
\usepackage[authoryear]{natbib}
\usepackage{todonotes}
\usepackage{fancyvrb}
\usepackage{pmboxdraw}
\usepackage[utf8]{inputenc}

\def\tsc#1{\csdef{#1}{\textsc{\lowercase{#1}}\xspace}}
\tsc{WGM}
\tsc{QE}
\tsc{EP}
\tsc{PMS}
\tsc{BEC}
\tsc{DE}
\begin{document}
\let\WriteBookmarks\relax
\def\floatpagepagefraction{1}
\def\textpagefraction{.001}

\shorttitle{Crowd-sensed sustainable indoor location services}

\shortauthors{Nasios et~al.}

\title [mode = title]{Citizen participation: crowd-sensed sustainable indoor location services}            



\author[1]{I. Nasios}[orcid=0000-0002-1765-0646]

\ead{ioannis.nasios@nodalpoint.com}

\author[2]{K. Vogklis}[orcid=0000-0002-3175-6658]
\ead{voglis@ionio.gr}

\author[3]{A. Malhi}
\ead{amalhi@bournemouth.ac.uk}

\author[4,3]{A Vayona}[orcid=0000-0002-3657-3826]
\ead{a.vayona@pontsbschool.com}

\author[5]{P. Chatziadam}[orcid=0000-0003-0501-5845]
\ead{panosc@ics.forth.gr}

\author[3]{V Katos}[orcid=0000-0001-6132-3004]
\ead{vkatos@bournemouth.ac.uk}

\affiliation[1]{organization={Nodal Point Systems},
    addressline={Mitropoleos 43}, 
    city={Athens},
    postcode={105 56}, 
    country={Greece}}

\affiliation[2]{organization={Ionian University},
    addressline={Ioannou Theotoki 72}, 
    city={Corfu},
    postcode={491 00}, 
    country={Greece}}

\affiliation[3]{organization={Bournemouth University},
    addressline={Fern Barrow}, 
    city={Poole},
    postcode={BH12 5BB}, 
    country={UK}}

\affiliation[4]{organization={Ecole des Ponts Business School, ParisTech},
    addressline={11 Rue Pierre et Marie Curie}, 
    city={Paris},
    postcode={75005}, 
    country={France}}

\affiliation[5]{organization={Foundation for Research and Technology – Hellas (FORTH)},
    addressline={Nikolaou Plastira 100, Vassilika Vouton}, 
    city={Heraklion},
    postcode={70013}, 
    country={Greece}}

\begin{abstract}
In the present era of sustainable innovation, the circular economy paradigm dictates the optimal use and exploitation of existing finite resources. At the same time, the transition to smart infrastructures requires considerable investment in capital, resources and people. In this work, we present a general machine learning approach for offering indoor location awareness without the need to invest in additional and specialised hardware. We explore use cases where visitors equipped with their smart phone would interact with the available WiFi infrastructure to estimate their location, since the indoor requirement poses a limitation to standard GPS solutions. Results have shown that the proposed approach achieves a less than 2m accuracy and the model is resilient even in the case where a substantial number of BSSIDs are dropped.

\end{abstract}




\maketitle

\section{Introduction and motivation}

There is a systematic and continuous growth of research in the domains of smart technologies and sustainability, exploring the overlapping and complementary agendas and use cases. However, there is limited research on sustainable integration and deployment of smart technologies in prominent research themes, such as the circular economy paradigm \citep{kirchherr2017conceptualizing}. The circular economy advocates the optimal utilisation of finite resources through a series of the so-called "R" practices - namely Reuse, Recycle, Repurpose, Rethink, Refurbish, Reduce, Renew, Repair, and Redesign, and this list keeps on growing \citep{potting2017}. Data-driven circular economy advocates the use of ICT to offer practical solutions to deliver a circular economy agenda. More specifically, data-driven Circular Economy (CE) is ``the utilization of reactive, adaptive, autonomous or collaborative objects and systems for economic and environmental value creation'' \citep{LANGLEY2021853}. The accelerated technological development cannot meet sustainable development goals if these are not explicitly defined and there is no elaborate sustainability strategy on how to meet these goals. In fact, there are several cases where technological advancements have been shown to have adverse effects on sustainability even in the cases of the technological solution aiming to solve a sustainability-efficiency challenge, as captured by the so-called Jevons' paradox \citep{POLIMENI2006344}. To avoid this, \cite{su10030639} argue that the Circular Economy "R"s should be considered when introducing digital technologies in the CE domain. 

This paper aims to introduce a comprehensive framework to showcase how human/citizen engagement can balance the need to invest in additional (hardware) infrastructure in order to offer smart services. Given a point or geographic region of interest, we distinguish between the role of a local citizen and visitor and show how the social capital can leverage technological enablers to offer a solution that respects the CE paradigm. We provide a fully working solution for estimating an indoor location, where the higher degree of citizen engagement, the higher location accuracy experienced by a visitor, without this resulting into an increase in a demand for hardware assets. 

The paper focuses on the intersection of social computing, social capital, participatory (or crowd) sensing and machine learning. It illustrates how a selection of the circular economy "R"s can be put to practice and develop solutions utilising existing technologies and infrastructure. By applying \textit{Rethink} and \textit{Repurpose} in an indoor location use case, research shows how to inject \textit{smartness} into a building by repurposing existing wifi signals and avoiding the introduction of additional specialised hardware. Increasing the utilisation of existing resources is a key milestone to change of mindset and sustainable thinking.    

The two main technological enablers contributing to this research are mobile crowd sensing (MCS) and machine learning (ML). MCS is a manifestation of participatory sensing \citep{burke2006participatory}. It is defined as \emph{"a new sensing paradigm that empowers ordinary citizens to contribute data sensed or generated from their mobile devices, aggregates and fuses the data in the cloud for crowd intelligence extraction and people-centric service delivery"} \citep{guo2014}. ML would allow the processing of data and the production of contextually useful and actionable information. Participants make use of their mobile devices that are already equipped with a plethora of sensors, to capture data surrounding their environment. This enables them to improve the quality of a specific range of services and needs through the application of ML. The use of crowds shows a number of benefits and advantages both in increasing the quality of the solution provided for a task \citep{Gong2018TaskAF}, as well as in offering more efficient and potentially sustainable options \citep{Du2020LetOC}.

 Using data collected by a large number of users, MCS services can provide more accurate and up-to-date information on indoor spaces, including the location of individuals and points of interest. This approach to indoor positioning and navigation is particularly useful in large, complex indoor environments such as airports, shopping malls, and hospitals, where traditional GPS-based positioning is often unreliable. In addition to improving the accuracy of indoor location services, MCS services also have the potential to promote sustainable development. By relying on crowd-sourced data from mobile devices, these services can reduce the need for dedicated indoor positioning infrastructure, which can be costly and environmentally unsustainable. Furthermore, by encouraging users to contribute data on their indoor movements and behaviors, these services can help optimize indoor spaces and reduce energy consumption, ultimately leading to more sustainable and efficient use of indoor environments.

However, subscribing to the common position that technology alone is not sufficient in achieving sustainability, we show how the social capital of a city can be recruited in providing a primitive function on a technical and operational level by both respecting the CE paradigm and encouraging the hospitality sentiment of a city. 

More specifically, from a technical and technological perspective we focus on indoor location-based services and propose a sustainable indoor positioning solution provided that there exists an adequate number of WiFi signals. In the context of this work, we consider sustainability to relate to the avoidance of introducing additional, purpose-specific hardware to offer indoor location services, which is traded off by social engagement through MCS activities. We also consider the case of minimising the need of frequent retraining the position estimators by the requirement of service resiliency, where it is expected that the positioning service will be operational even in the cases of losing a significant number of BSSIDs over a short amount of time.

The remainder of the paper is structured as follows. Section \ref{social} elaborates on the rationale for recruiting citizens or habitats of a particular location to contribute to computational activities that can be used towards a \emph{common good}. Section \ref{literature} presents an overview of state-of-the-art research in indoor positioning. Sections \ref{setup} and \ref{model} explain the MCS experimental arrangements and ML development and evaluation, respectively. Finally, Section \ref{conclusion} presents the concluding remarks.

\section{From social computing to actionable intelligence and sustainability}
\label{social}

One of the most prominent functions of user engagement is feedback and product or service reviews. The proliferation of online shopping enabled consumers to generate and publish reviews. Reputation systems can be found for almost any service provision or product purchase activity. From an information-centric perspective, the use of the \emph{crowd} manifests in the crowdsourcing and participatory sensing paradigms. Users become \emph{prosumers}, that is, both producers and consumers of information. Review services such as Tripadvisor for hotels, Google reviews and Trustpilot for companies, Checkatrade for trade people, etc., are examples of business models that directly capitalize on capturing and leveraging the added value of the crowd \citep{mariani2021innovation} - in this case, the consumers of the relevant product or service. On a second tier, platforms ingesting and analyzing reported visitors experience in a certain place can provide a more substantial picture of the \emph{collective sentiment} for a certain geographical region. Platforms such as barregi.carto.com, and insideairbnb.com analyze Airbnb review data and provide actionable and contextualised intelligence.

Social media provide a less structured, yet quite effective venue for location contextualisation and intelligence \citep{harrigan2015modelling}. Social media posts may contain relatively coarse-grained location information which is contextualised by a variety of information, ranging from historical, point of interest evidence, to personal and collective views, moments, events, and emotions \citep{park2020visualizing,xu2017building}. In a city context, the habitats and visitors of a smart city assume the prosumer role \citep{faysal2019} by interacting with the city's assets in order to enjoy and contribute to the city's services. Citizens are effectively becoming information processing nodes and the city's infrastructure provides the synapses to enable information flows and allow transfer of value between the physical and cyber plane \citep{vrana2021cyber}. Citizens interact with the digital services through smartphones and this is captured also in the digital service providers' response who invest in developing suitable smartphone apps. 

In addition to the data captured in (near) real-time in the aforementioned platforms, there have also been a number of studies evaluating cities over a number of features and indicators. The Cities in Motion Index \citep{cim2019} maintains 96 variables over 9 indicator categories, including population, human capital, and technology. \cite{imd2019} offer a digital competitiveness ranking for 63 cities, used to explore how digital technologies can contribute to economic and social transformation. The C40 project started as an initiative capturing use cases of 40 cities adopting a circular economy agenda and has lately grown to encompass over 100 cities \citep{C40}.

\begin{figure*}[h]
    \centering
    \includegraphics[width=0.8\textwidth]{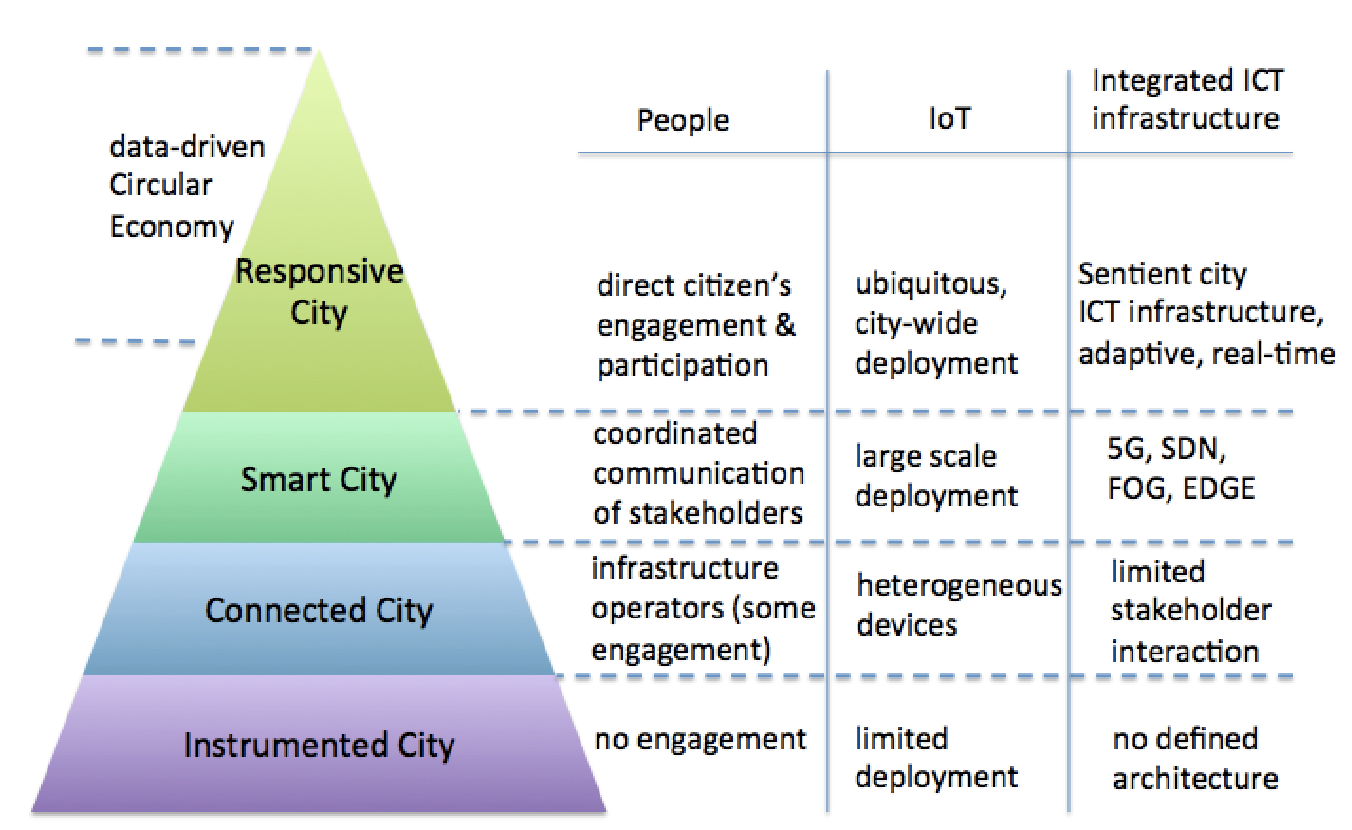}
    \caption{Smart city maturity model \citep{ideal2019}}
    \label{fig:smart_maturity}
\end{figure*}

The data and features presented above are a glimpse of the wealth of studies and metrics available for assessing a city. When considering the journey of a city in becoming smart, maturity models can be employed. A smart city maturity model evaluates the state of a city against a selection of dimensions and can illuminate the paths of a strategic roadmap for the city to achieve a higher level of \emph{smartness}. This research subscribes to the smart city maturity model defined in the Ideal Cities project (Figure \ref{fig:smart_maturity}). According to this model, there are four distinct levels, namely the \textbf{Instrumented City}, the \textbf{Connected City}, the \textbf{Smart City} and the \textbf{Responsive City} \citep{ideal2019}. While the lowest level of the Instrumented City refers to just capturing data from utilities sensors and reporting them to the respective utility provider (and perhaps the local authority), at the highest end of the model, the Responsive City, all assets, including the citizens and visitors, are in complete sync and aware of their state and presence of their peers and other resources. 

Big data analytics capabilities are fundamental in achieving the Smart City status. The inclusion of intelligent assets \citep{morlet2016intelligent, rossi2020circular}, supported by the continuous, real-time flow of information, combined with the actuator capabilities of IoT assets can repurpose say a park to an overflow parking lot during a sports event or concert, or turn a parking lot into a vaccination centre when handling a pandemic. The Responsive City is the maturity level at which sustainability can be achieved \citep{ideal2019}.

As depicted in Figure \ref{fig:smart_maturity}, the dimensions of a smart city consist of \emph{infrastructure}, \emph{IoT technologies} and \emph{People}. While more dimensions can be found in the recent literature \citep{itu2019,gasco2018building}, we highlight these three as relevant to this study. With regards to the People dimension in particular, it can be seen that the city can reach the Responsive level only if there is direct citizen's engagement and participation. 

When it comes to citizen participation on unpaid activities, such as crowdsourcing, fundraising events, community work and so forth, cultural and societal aspects are a key consideration. There are several psychological factors that influence one’s decision to participate in fundraising events, such as: social identity and sense of belonging sense of belonging, altruism, a personal connection, but also peer social pressure. The benefits and advantages of the contribution of social capital and social cohesion have been acknowledged and witnessed in several instances. In \cite{jarvis2022} a regression analysis of Cities in Motion factors \citep{cim2019} and computing devices with software vulnerabilities from Shodan\footnote{\url{https://shodan.io}} revealed that there is a significant relationship between cybersecurity exposure, social cohesion and human capital of a city. \cite{ertio2017citizens} elaborate on how participatory platforms can actively organise through their own networks and communities and promote the concept of citizen-initiated planning.

A scenario that combines hospitality and healthcare using the power of the crowd leverages the so-called Nearest Doctor Problem \citep{drosatos2014efficient}. In busy tourist areas or during events that attract many visitors, local health services can become overwhelmed. The Nearest Doctor concept involves quickly identifying a qualified doctor near the incident who is willing to attend to it. However, since the doctor may not be available due to being on vacation, their privacy and availability must be taken into account. In order to maintain privacy, the doctor can have an agent on their mobile device that will respond only if the doctor is available. Once the doctor is available, a privacy calculation can be done to reveal the location and identity of the closest doctor, while keeping the location and identity information of all other doctors private \citep{ideal2019}.

When discussing minimization of building energy consumption it is well established  \citep{KANG2022104193} that understanding the movement of occupants inside buildings is crucial for building design and operations. The distribution of people on walkways and halls must be considered, and citizen-initiated location services appear as a promising alternative. 

To cope with the dynamic nature of space planning the authors in~\citep{UGAIL2021102791} proposed a social distancing enhanced automated optimal design for indoor spaces. Given the dimensions and the floor plan of a given physical space, an optimization problem is solved that corresponds to the requirements on the social distancing criteria between people and the imposed constraints such as the position of doors, windows, and walkways.

Gamification in crowd-sourcing is another area of research that has received considerable attention \citep{morschheuser2016gamification, sigala2015gamification}. The development of the popular Pokémon Go augmented reality game and its use of real-world locations is a representative example \citep{carbonell2018pokemon}. The game's challenges requiring users to visit selected points of interest led to an increase in foot traffic in certain areas, as players flocked to locations with a high density of Pokémon Go artefacts. Google Maps was able to use these data to improve its way-finding and routing capabilities, as it was able to use more accurate data on foot traffic and popular destinations. In addition to the benefits of sustainability and cost effectiveness for Google Maps itself, ``unintended'' benefits were observed for players who were incentivized to engage in physical activity and explore their local communities \citep{althoff2016influence, chong2018going}.

This paper proposes an approach where users can contribute to offering the primitive function of location services through MCS annotation. This approach can be fully implemented only when local residents are regularly engaged in technology through their smartphones. Apart from the sustainability goal this approach offers, it can also be used as an indicator of the locals' commitment and attitude towards both hospitality and sustainability.

\section{Location based services}
\label{literature}
Location-based services (LBS) have received the attention of researchers due to their practical benefits supporting a wide range of use cases and relatively ease of implementation on different portable smart devices \citep{huang2018location}. LBSs have numerous benefits, including increased convenience, enhanced safety, improved people engagement, and have become an integral part of the mobile experience. One of the most representative applications is location-based advertising and marketing~\citep{lbs_a1,lbs_a2}. This type of LBS provides users with targeted advertising or promotions based on their current location. For example, a user who walks past a coffee shop may receive a coupon for a discount on a cup of coffee. Another common application is Location-based social networking~\citep{lbs_b1,lbs_b2}, which provide services to their users, taking into account both spatial and social aspects.  Examples of location-based social networking services include Foursquare~\footnote{\url{https://foursquare.com/}}, Swarm~\footnote{\url{https://www.swarmapp.com/}}, and Facebook Places~\footnote{\url{https://www.facebook.com/places/}}. The most generic use case is location-based information services~\citep{lbs_c1}. This type of LBS provides users with information that is relevant to their current location. Examples include turn-by-turn navigation, weather alerts, and points of interest, such as restaurants or gas stations.

Most of these applications are dependent on the accuracy of location information on smart devices. As GPS cannot provide reliable indoor navigation services, different methods such as FM, RFID, ultra-wideband and others have currently been deployed \citep{shi2016survey}. 

Indoor fingerprint location, which utilises a set of precalibrated RSS fingerprints, can be categorized into three methods based on the estimation approach used, namely probabilistic, deterministic, and machine learning. The probabilistic approach uses Bayesian inference for computation of each grid point's probability and estimating the position of the user. The deterministic approach divides the environment into cells for creating a radio map to obtain the position estimate, which is determined by the most accurate match between the radio map's measurements and any new measurement. The machine learning approach uses a series of algorithms such as K-Nearest Neighbor, Random Forest, K-means, Support Vector Machine, and Artificial Neural Network algorithms. Due to their wide deployment, WiFi enabled systems are most commonly used for indoor positioning systems. The most popular techniques used for WiFi based location services are based on fingerprinting or trilateration. However, when channel state information (CSI)-based fingerprinting is used to attain centimeter-level localisation accuracy \citep{nessa2020survey}, advanced network interface cards \citep{wang2016csi} are needed for measuring CSI, adding to the cost. 

Classifiers are the most commonly used algorithms in localization to extract key features from signals. Clustering is performed based on these retrieved features using the fingerprint approach. Feature extraction is a crucial step for the identification and mitigation of Non Line Of Sight (NLOS) features \citep{fan2019non}. Overfitting and data mapping are important issues with fingerprint-based localization systems. Of the classification algorithms, SVM is more effective because it models linear and nonlinear relations with better generalization. It utilises kernels for detection of the difference between two points that belong to two separate classes. A disadvantage of using SVM is that if the number of support vectors is large, the SVM-based method takes longer time and utilizes a significant amount of memory. Indoor localization based on Decision Trees outperforms other classification methods such as K-NN and Neural Networks in terms of improving localisation accuracy as reported by \cite{nessa2020survey}.

 In recent years, Visible Light Communication (VLC) based indoor positioning has shown significant improvements which include high durability, low cost, and environmental friendliness. In such systems, the transmitter utilizes existing indoor lighting systems for the emission of optical signals, which may contain their positioning coordinates or ID. There is an increasing trend of utilizing machine learning methods to achieve better accuracy and robustness in the VLC-based indoor positioning systems. \cite{van2017weighted} show how WK-NN and KNN can be used to achieve a high degree of accuracy which gives about 50\% improvement over conventional RSS-based trilateration. In another research by \cite{saadi2016led}, a two-level clustering method based on RSS values and position coordinates is proposed to determine several cluster centers. The final position is calculated by minimizing the Euclidean error of the given LED RSS values with that of cluster centroids. 
 
 The work of \cite{sthapit2018bluetooth} proposes a low energy-based bluetooth-based machine learning location and tracking system for indoor positioning. The average estimation was only 50 cm with the proposed method.

\cite{link2011footpath} propose an indoor navigation system that compares First-fit and Best-fit algorithms. The analysis was carried out primarily using the accelerometer and compass that are available in every smartphone, correlated with OpenStreetMap data feeds. An analysis of this study suggests that the first fit is less accurate around elevators and metal structures as the compass performs distorted measurements due to the proximity of the metal structures. As the best fit constantly updates penalty points, it provides better accuracy in comparison to the first fit. \cite{megalingam2013larn} propose the Location Aware and Remembering Navigation (LARN) algorithm which is based on calculating the shortest distance to various points using Dijkstra’s shortest path algorithm. LARN is particularly designed for applications that support wheelchair users. The algorithm requires mapping the initial floor plan and dividing the same into grids of equal size. Each grid is then assigned to a unique address. Once the component running the LARN algorithm finds the shortest path, it is passed over to the microcontroller that controls the movement of the wheelchair. One of the major advantages of LARN is that it does not require a wireless device.

In \cite{SHAH2018298} the authors propose an indoor location tracking method utilizing the received Signal Strength Indicator (RSSI) from various sensors. The proposed localization method is based on the 3D weighted centroid algorithm and does not need any additional equipment.

The aforementioned state-of-the-art research considers technologies for indoor location tracking without explicit considerations towards sustainable computing, and circular economy principles. This paper illustrates how key enablers of data-driven circular economy can be put into practice by repurposing the existing WiFi infrastructure to develop indoor location tracking using machine learning approach.
\section{Experimental setup}
\label{setup}
\begin{figure*}[!h]
    \centering
    \includegraphics[width=0.8\textwidth,angle=0]{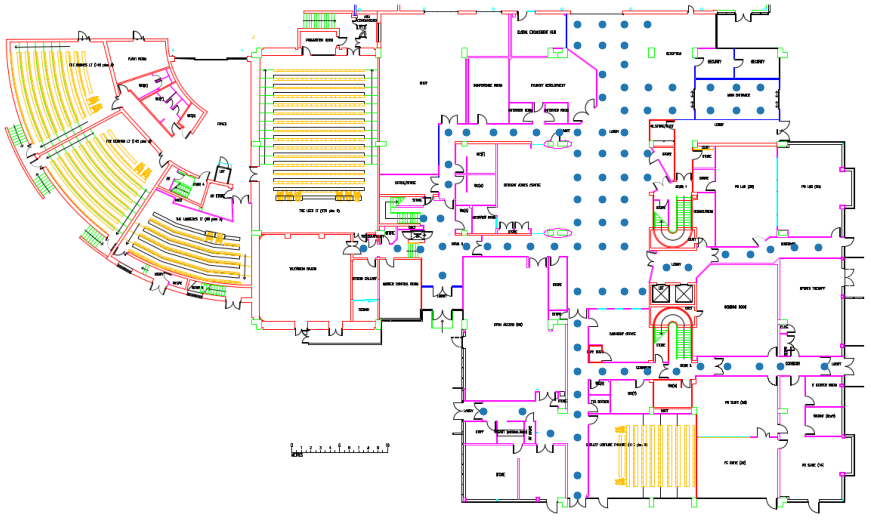}
    \caption{Ground Floor Map with Label Points}
    \label{fig:katopsi2Pointsv2nonumscrIm}
\end{figure*}
To address the technical challenges and to illustrate the principles of MCS, an experiment was conducted at indoor premises, employing the existing WiFi infrastructure of a building situated at Bournemouth University in the UK. This venue was selected as it provided ease of access to the ceiling that had already a grid of lights with an ideal to the experiment distances, as well as adequate spatial complexity (straight and looping corridors, and open spaces) and size. 

The first task was to establish the ground truth representation. This was achieved by annotating the ground floor map shown in Figure~\ref{fig:katopsi2Pointsv2nonumscrIm} with a grid of labeled points. Each labeled point corresponds to a set of indoor coordinates in meters where $(0, 0)$ was assigned to the bottom right corner of the building. In Figure~\ref{fig:katopsi2Pointsv2nonumscrIm}, we show approximately $100$ ground truth points that form a grid of approximately $2.5$ meters wide. In this way, we create a set of $N_{max}$ checkpoints $\mathcal{X} = \{ (x_i, y_i), \ \ \emph{for}\ x=1, \ldots, N_{max} \}$, which will represent known coordinates within the building in meters. 

In order to create the initial dataset, having defined the initial set of checkpoints, an Android data acquisition application was created and shared with the participants. The users were instructed to hold the mobile device under the label and tap on the corresponding number displayed on their screen. The data acquisition application included a connected graph of all labels in order to provide easy access to the users by presenting only the adjacent labels (nodes). When the user clicks on a label, the application records the specific label ID~\footnote{Also referred as checkpoint} with the current timestamp and the screen refreshes with the labels adjacent to the selected point.

\begin{figure}[!h]
    \centering
    \includegraphics[width=0.35\textwidth]{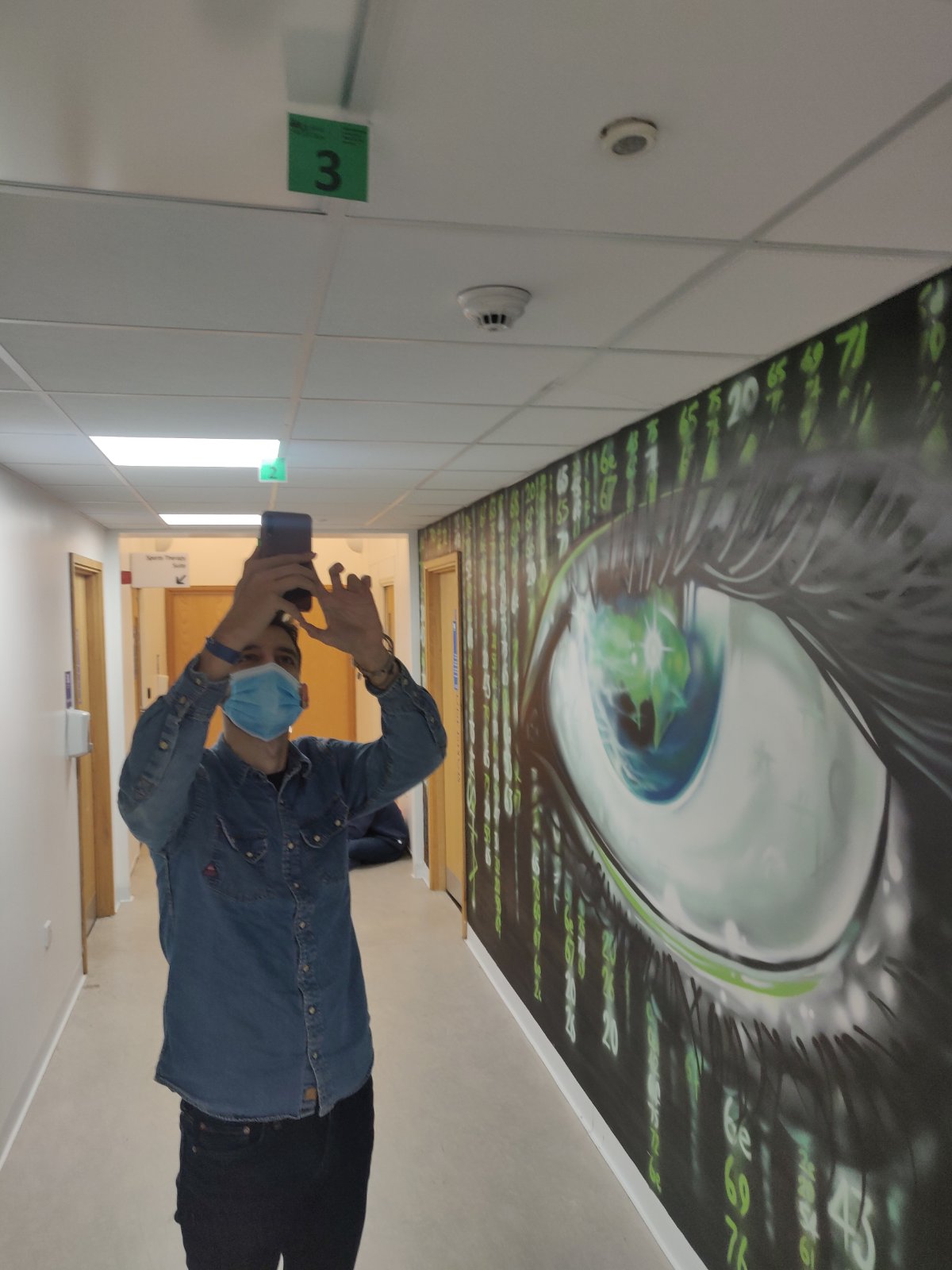}
    \caption{Citizen's engagement: data annotation}
    \label{fig:acquire}
\end{figure}

\begin{figure}[!h]
    \centering
     \scalebox{1}[-1]{\includegraphics[width=0.5\textwidth,angle=0]{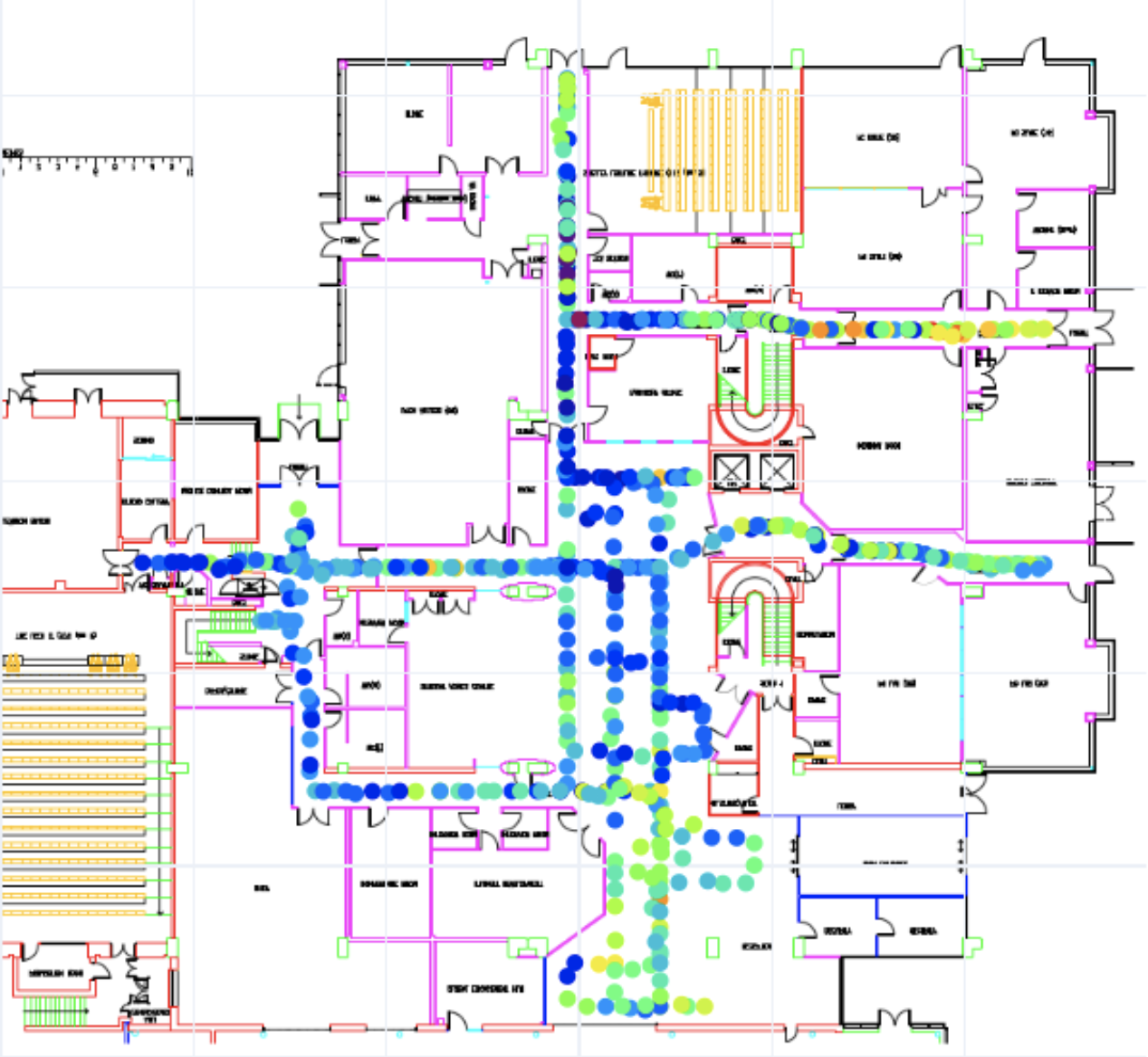}}
    \caption{Ground Floor Map with a sample path}
    \label{fig:sample_path}
\end{figure}

The above data collection protocol would allow a user to seamlessly navigate within the building and create predefined or ad-hoc paths. An example of user participation for data acquisition is shown in Figure~\ref{fig:acquire} and the sample path is presented in Figure~\ref{fig:sample_path}. We can extend the data gathering protocol with ideas of gamification and ludrification to create a more fun experience for the user/citizen. 

The raw data from the application are captured and stored in a series of text log files, where each file represents a unique recorded path. The user can decide when to begin or end a specific path. In addition to user input at the labeled points, the application records a list of surrounding WiFi BSSIDs~\footnote{BSSID stands for Basic Service Set Identifier. This is the MAC physical address of the access point or wireless router that is used to connect to the WiFi.} along with its signal strength in dB every 2 seconds.In addition, sensor data from mobile device, including gyroscope, accelerometer, and magnetometer readings are recorded every $50$ milliseconds. These are included in the published dataset to enable future research on dead reckoning.

The dataset is annotated by associating the time-stamped WiFi data with a specific checkpoint. In the following snippet, for example, the associated WiFi data for Checkpoint $12$,  pressed at {\tt 1633437608241}, (i.e. {\tt 12:40:08.241}): 

\begin{Verbatim}[fontsize=\footnotesize, frame=lines, label=Raw data file around label point 12,commandchars=\\\{\}]
1633437600157 TYPE_GYROSCOPE 0.118198946 -0.20234315 -6.8720314E-4  3
1633437600157 TYPE_ROTATION_VECTOR 0.103371434 -0.13861331 -0.6763132  3
1633437600157 TYPE_ACCELEROMETER 1.1192428 3.6858442 9.081061  3
1633437600157 TYPE_MAGNETIC_FIELD -23.6625 22.96875 165.76875  0
...
\fbox{1633437608241 TYPE_CHECKPOINT 12}
...
1633437608489 TYPE_GYROSCOPE -0.06230642 0.028251685 0.22265382  3
1633437608489 TYPE_ROTATION_VECTOR 0.33039182 -0.44274282 -0.6228367  3
1633437608489 TYPE_ACCELEROMETER -0.13519448 8.687142 4.922335  3
1633437608489 TYPE_MAGNETIC_FIELD -29.7 -0.31875002 183.45001  0
...
1633437608615 TYPE_WIFI f4:db:e6:aa:bb:cc SSID1 -57 5620  
1633437608615 TYPE_WIFI f4:db:e6:aa:bb:dd SSID2 -57 5620  
1633437608615 TYPE_WIFI f4:db:e6:aa:bb:ee SSID3 -57 5620  
...
\fbox{1633437611901 TYPE_CHECKPOINT 13}
...
\end{Verbatim}

\section{Machine Learning Modeling}
\label{model}
The aim of the proposed approach is to create/train a regression model that would take s list of nearby WiFi BSSIDs along with the corresponding signal strength levels as input and will approximate the internal coordinates $(x_i, y_i)$ at the predefined checkpoints $\mathcal{X} = \{ (x_i, y_i), \ \ \emph{for}\ x=1, \ldots, N_{max} \}$. This is a two-step process: a) first, the data are prepared and expressed in an appropriate tabular/numeric format and b) a suitable parametric regression model is estimated through training.  

To suit the participatory concepts of the proposed approach, the training procedure must be able to incorporate new data when they are available (online training) and in order to continuously improve - or perhaps avoid any detrimental effects to - the predictions.

\subsection{Data pre-processing}
During the citizen participation phase, the routes are captured using a mobile device and stored in the form of a text log file, one for each distinct route. Routes that contain more than $4$ checkpoints are only saved and considered in order to avoid really short and probably erroneous paths. Then all available WiFi present signals across all paths are aggregated to calculate the frequency of appearance of each and every BSSID. To keep one BSSID as input element for the modeling phase, it is required that this BSSID be present at at least $50$ checkpoints. This makes the approach more robust as it allows the preprocessor to discard remote WiFi antennas from other buildings that are visible in a limited area. Furthermore, WiFi SSIDs containing smartphone model keywords such as \emph{Android, Galaxy, iPhone, HUAWEI, Pixel, Nokia, Honor} were also discarded, as they correspond to mobile hotspots. These SSIDS are not fit for purpose and highly volatile, as they are expected to change position or disappear, causing deterioration to the predictions.

\begin{figure}[!h]
    \centering
    \begin{tikzpicture}
    \node[anchor=south west,inner sep=0] at (0,0) {\includegraphics[width=0.48\textwidth]{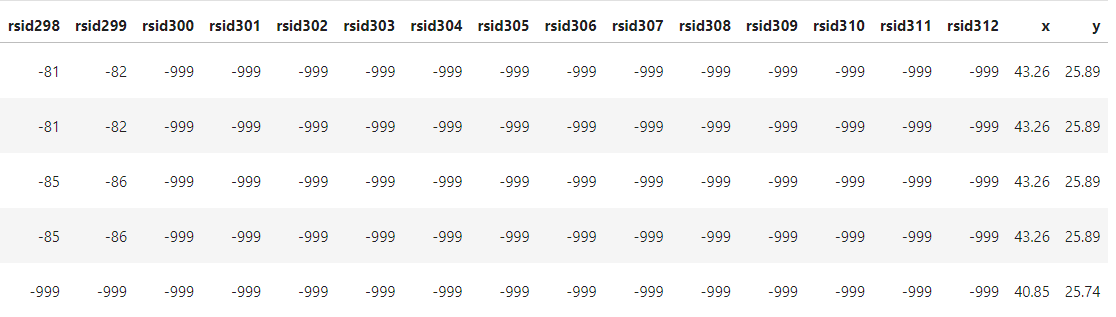}};
    \draw[red,ultra thick,rounded corners] (0,0.5) rectangle (1.2,2.0);
\end{tikzpicture}
    
    \caption{Data sample of the last columns}
    \label{fig:data_part2Im}
\end{figure}

A crucial training parameter is the number of unique BSSIDs collected from all paths. In the experimental testbed $N_{bssid} = 312$ unique BSSIDs were recorded. Each row of the dataset corresponds to a checkpoint and contains the $(x, y)$ position in meters to be used as a target feature. The input vector is $N_{bssid}$-dimensional and contains the signal strength of visible WiFis around the specific checkpoint or a value representing an unknown set to $-999$ otherwise. Figure~\ref{fig:data_part2Im} presents a sample of five data rows where it can be seen that {\tt rssid298} and {\tt rssid299} are visible from the first $4$ checkpoints and have signal strengths around $-82$ dB, while the remaining WiFis from {\tt rssid300} to {\tt rssid312} are not visible at all.  

The initial dataset consists of $14373$ rows that correspond to $423$ unique paths. Each data row contains a timestamp indicating when this sample was recorded and the path name to which this sample belongs. This yielded $423$ unique paths with an average of approximately $34$ rows - timestamps per path.

\begin{figure}[h]p
    \centering
    \includegraphics[width=0.48\textwidth,trim=0 0 0 17, clip]{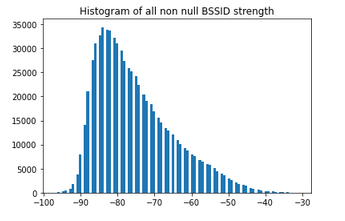}
    \caption{WiFi strength distribution}
    \label{fig:wifi_strengthsIm}
\end{figure}

Figure \ref{fig:wifi_strengthsIm} presents a histogram of all signal strengths captured during the course of the experiment. Signal strengths have a range of $-97$ to $-31$, with a median of $-78$ and an average of $-75.4$. The right-skewed distribution indicates that most of the time, most of the reached WiFi signals are not as powerful. This is expected, as it is well known that most WiFis can easily fall out of reach at any time due to the inherent WiFi range limitations.

\begin{figure}[h]
    \centering
    \includegraphics[width=0.48\textwidth,trim=0 0 0 20, clip]{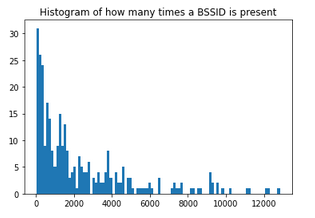}
    \caption{WiFi BSSID frequency of appearances}
    \label{fig:bssids_presenceIm}
\end{figure}

Some of the WiFis were detectable only $46$ times ($0.3\%$) and some others $12823$ times ($89.2\%$) with a median of $1300$ ($9\%$). In Figure \ref{fig:bssids_presenceIm} the frequency distribution of WiFi appearances in the dataset is presented.

\subsection{Modeling experimental setup}
The initial investigation involved three diverse regression models: (a) gradient boosting trees  (\textbf{GBM}) referred as LightGBM~\citep{gbm1,gbm2,gbm3} (b) recurrent neural network (\textbf{RNN}) using Keras - tensorflow implementation \citep{nn1, nn2} and (c) k-nearest neighbors (\textbf{KNN}) which is an unsupervised approach. All three models, despite adopting considerably different approaches, have been accepted to be suitable for this task, as evidenced from the best solutions in a relevant machine learning competition~\citep{Kaggle}.

The evaluation protocol is \emph{5-fold cross validation} and Folds are \emph{split by path} so that a complete path belongs to a single fold. For comparison reasons, the split was kept the same for all experiments. The optimized metric, is the \emph{mean position error} that is, the Euclidean distance in meters:
\begin{equation}
    \emph{MPE} = \sum_{i=1}^{N_{max}} \sqrt{\left( x_i - \tilde{x}_i\right)^2 + \left( y_i - \tilde{y}_i\right)^2}), 
\end{equation}
\noindent where: 
\begin{itemize}
\item [$N_{max}$] is the number of rows in the dataset \\
\item [$(x_i, y_i)$] are the ground truth locations for a given test row\\
\item [$(\tilde{x}_i, \tilde{y}_i)$] are the predicted locations for a given test row
\end{itemize}

\subsection{Experimental evaluation}
The experimental investigation initially involves a comparison of the multiplicity of errors (MPE) of each of the selected regression models. The parameters for \textbf{GBM} are shown in Table~\ref{tab:gbm_params}. For the implementation of the recurrent neural network, LSTM~\citep{lstm1,lstm2} layers were used with the architecture described in Table~\ref{tab:nn}. The simple \textbf{KNN} model consists of a single parameter set to $k=45$.

\begin{table}[width=.9\linewidth,cols=2,pos=h]
\caption{GBM model parameters}\label{tab:gbm_params}
\begin{tabular*}{\tblwidth}{@{} LL@{} }
\toprule
Parameter & GBM \\
\midrule
boosting\_type            & gbdt      \\ 
 objective                 &  regression  \\
 metric                     & l1, l2 \\
learning\_rate  & 0.005 \\
feature\_fraction & 0.9 \\
bagging\_fraction &  0.7 \\
bagging\_freq &  10 \\
max\_depth & 8 \\
num\_leaves & 128   \\
max\_bin & 512 \\
num\_iterations & 15000 \\
\bottomrule
\end{tabular*}
\end{table}

\begin{table}[width=.9\linewidth,cols=3,pos=h]
\caption{RNN architecture with $2$ Dense and $3$ LSTM Layers}\label{tab:nn}
\begin{tabular*}{\tblwidth}{@{} LLL@{} }
\toprule
 Layer (type) & Output shape  & \# params \\
\midrule
 dn1 (Dense) & (None, 256) & 80,128 \\
 bn2 (BatchNormalization)& (None, 256) & 1,024 \\     
 do2 (Dropout) & (None, 256) & 0.3 \\
 dn2 (Dense) & (None, 256) & 65,792   \\  
 resh (Reshape) & (None, 1, 256) & 0.2 \\       
 bn3 (BatchNormalization)& (None, 1, 256) & 1,024 \\     
 lstm1 (LSTM) & (None, 1, 128) & 197,120 \\
 do3 (Dropout) & (None, 1, 128) & 0.2 \\   
 bn4 (BatchNormalization) & (None, 1, 128) & 512 \\      
 lstm2 (LSTM) & (None, 1, 64) & 49,408 \\
 do4 (Dropout) & (None, 1, 64) & 0.1  \\    
 bn5 (BatchNormalization) & (None, 1, 64) & 256 \\      
 lstm3 (LSTM) & (None, 32) & 12,416 \\
 bn6 (BatchNormalization) & (None, 32) & 128 \\      
 xy (Dense) & (None, 2) & 66 \\
\midrule
Total params: 409,122 & & \\
Trainable params: 407,026 & & \\ 
Non-trainable params: 2,096 & & \\
\bottomrule
\end{tabular*}
\end{table}

As depicted in Table \ref{table:model_performance}, \textbf{GBM} performs better than the other two models, followed by \textbf{RNN}. Furthermore, merging the output of these models in a $2:1$ ratio reduces the mean position error. Lastly, \textbf{GBM} exhibits the smallest standard deviation of MPE across 5-folds, which is a desirable performance indicator.

\begin{table}[width=.9\linewidth,cols=3,pos=h]
\caption{Mean Position Error of different models}
\label{table:model_performance}
\begin{tabular*}{\tblwidth}{@{} LCC@{} }
\toprule
 Model & Mean MPE  & Std MPE\\ 
 \midrule
 GBM & 1.95  & 1.53 \\
 RNN & 2.08 & 1.69 \\
 KNN & 2.40  & 1.76 \\  
\midrule 
best ensemble & 1.92 & 1.53 \\
\bottomrule
\end{tabular*}
\end{table}

\begin{figure}[h]
    \centering
    \includegraphics[width=0.48\textwidth,trim=20 20 0 20, clip]{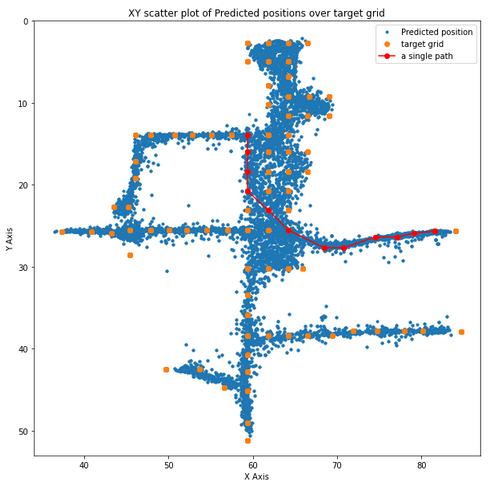}
    \caption{GBM predictions over target grid}
    \label{fig:fullgrid_preds_and_apathIm}
\end{figure}

In Figure~\ref{fig:fullgrid_preds_and_apathIm} the \textbf{GBM}'s $(x, y)$ predictions are represented as blue points and the ground truth as orange points. The macro structures (hallways and rooms) within the building are clearly visible. A path of a single participant (ground truth) is depicted with red connected points. It is visually evident that the \textbf{GBM} model manages to generalize and create meaningful predictions even between successive ground truth points. Some sparse outliers exist, but they can be easily filtered out by a smoothing process. 

\begin{figure}[h]
    \centering
    \includegraphics[width=0.48\textwidth, trim=0 0 0 15, clip]{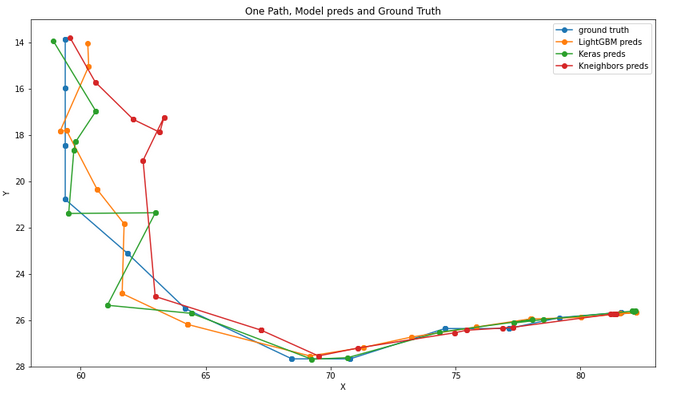}
    \caption{Models Predictions of one random Path}
    \label{fig:onePathRIm}
\end{figure}

An example of the predictions of different models for one single path is shown in Figure \ref{fig:onePathRIm}. The same path is also marked, with a red line, in Figure \ref{fig:fullgrid_preds_and_apathIm}. For this path, the mean errors are: 
\emph{1.25m} for gradient boosting machine,
\emph{1.52m} for Neural Network and
\emph{1.91m} for nearest neighbors model. At the left side of the plot, predictions for all models are further away from the true path, while at the right part of the plot, predictions are forming a path that matches with the true path. From this plot alone, it is easy to distinguish a hallway (right) from a larger room or hallway (left).

In the remaining subsections, a set of experiments measuring how different conditions may affect the MPE metric is presented. The evaluation approach considers the following two out of the three smart city dimensions against the data-driven elements as follows:

\begin{itemize}
    \item \textbf{People:} path data. How much data (in the form of paths) should be collected for the prediction error to be sufficiently small? What is the expected effort that citizens would need to invest in order to obtain a viable indoor location service? 
    \item \textbf{Infrastructure:} WiFi/BSSID data. What is the minimum number of BSSID points required to train the model? To what extent does the size of the grid affects the accuracy? Also, what is the robustness of a trained model when BSSIDs are dropped during prediction?
\end{itemize}

The IoT dimension is not considered at this stage of the evaluation, as it has been trivially addressed earlier (the removal of BSSIDs associated with smart phone makes and models). In addition to the evaluation approaches introduced above, a comparison of the influence between the people vs. the infrastructure originating data is also performed, in order to understand the key factors affecting the performance of the model.

\subsubsection{Citizen participation: Impact of the size of the data set on overall accuracy}
In order to establish the number of sufficient paths needed to achieve an acceptable level of accuracy, a training procedure (5-folds) was repeated using increasing subsets from the total of $423$ paths. A random sample of a fixed percentage from the complete set of paths (75\%, 50\%, 25\%) was picked. As some paths may be more prevalent than others, the experiment was run multiple times for every model, and each time, a different set of paths were selected. The results are summarised in Table \ref{table:model_performance_less_paths}.
As expected, the more paths included, the better the performance. The reduction in mean MPE seems to be in the order of centimeters after a certain number of paths was collected. Looking at these results, we conjecture that there exists a point where adding more participants will not affect the overall MPE accuracy, and this can be a topic for future investigation.

\begin{table}[width=.9\linewidth,cols=4,pos=h]
\caption{Performance of different models with increasing number of participants (paths) }\label{table:model_performance_less_paths}
\begin{tabular*}{\tblwidth}{@{} LCCC@{} }
\toprule
 Model & Paths & Repeats  & Mean MPE \\
 \midrule
 GBM & 106 & 50 & 2.21   \\
 RNN & 106 & 50 & 2.40  \\
 KNN & 106 & 50 & 2.62   \\  
 \midrule
 GBM & 211 & 40 & 2.06   \\
 RNN & 211 & 40 & 2.25  \\
 KNN & 211 & 40 & 2.48  \\  
\midrule
 GBM & 317 & 30 & 1.99  \\
 RNN & 317 & 30 & 2.15  \\
 KNN & 317 & 30 & 2.43  \\  
 \midrule
 GBM & 423 & 1 & 1.95  \\
 RNN & 423 & 1 & 2.08 \\
 KNN & 423 & 1 & 2.40 \\  
 \bottomrule
\end{tabular*}
\end{table}

\begin{figure}[h]
    \centering
    \includegraphics[width=0.5\textwidth, trim=40 40 0 40, clip]{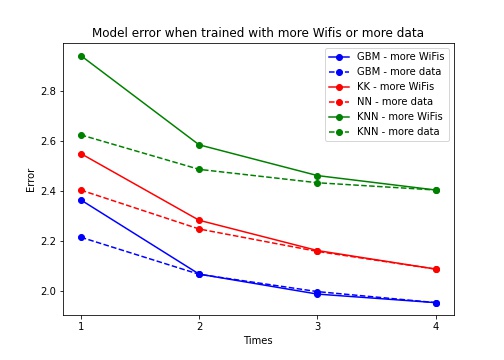}
    \caption{Models performance for increasing number of participants (paths) MHPWS: Comparison of models performance on increased WiFis or data}
    \label{fig:models_errorsIm}
\end{figure}

\subsubsection{Impact of number of different BSSIDs on accuracy}
The experiment involved the collection of 312 different Wifi BSSIDs. By removing a fixed percentage of BSSIDs and simulating conditions where some BSSID signals never existed, we can further measure the performance of the models. 

We used 75\%, 50\% and 25\% of the 312 unique WiFi BSSIDs (234, 156 and 78 WiFis). As some WiFis may be more important than others, we ran our experiments multiple times for every model and each time, a different subset of WIFis was selected in random. The fewer the available WiFis, the more experiment repetitions in order to obtain more accurate results as the standard deviation increased. The results are shown in Table \ref{table:model_performance_less_wifis}.

As expected, the higher the availability of WiFi antennas, the better the performance of the models. The results could have been improved if there had been even more WiFi in place. However, inspecting the trend between the available Wifis and the score, it is argued that the actual performance is too far from saturation.

 Figure~\ref{fig:models_errorsIm} shows the performance increase when introducing WiFi points and is compared with that of more data (paths). The trade-off is apparent and shows the preference of having more WiFis over path data.

\begin{table}[width=.9\linewidth,cols=5,pos=h]
\caption{Performance of different model with less WiFi antennas installed}
\label{table:model_performance_less_wifis}
\begin{tabular*}{\tblwidth}{@{} LCCCC@{} }
\toprule
 Model & WiFis & Repeats  & Mean Error & Std \\ [1ex] 
 \midrule
 GBM & 234 & 30 & 1.98 & 0.01  \\
 RNN & 234 & 30 & 2.16 & 0.02 \\
 KNN & 234 & 30 & 2.46 & 0.02  \\  
 \midrule
 GBM & 156 & 40 & 2.06 & 0.03  \\
 RNN & 156 & 40 & 2.28 & 0.03 \\
 KNN & 156 & 40 & 2.58 & 0.04  \\  
 \midrule
 GBM & 78 & 50 & 2.36 & 0.08  \\
 RNN & 78 & 50 & 2.55 & 0.06 \\
 KNN & 78 & 50 & 2.94 & 0.10  \\  
 \bottomrule
\end{tabular*}
\end{table}

\subsubsection{Model Resilience}
As mentioned, the models were trained with 312 different WiFi BSSIDs in place. We now consider the case where some of the BSSIDs become unavailable due to single fault, replacement of BSSID (so the new values are not registered and included in the training prior to a new round of user participation), or bigger events such as a fire that may affect the BSSID operational map.

\begin{figure}[h]
    \centering
    \includegraphics[width=0.48\textwidth,trim=20 20 0 20, clip]{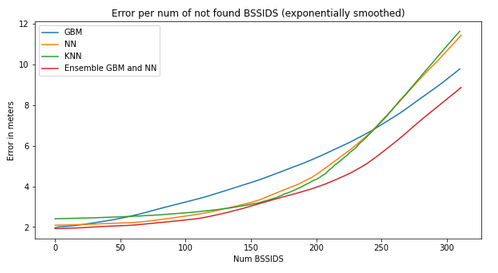}
    \caption{Model Performance plot over various missing WiFis}
    \label{fig:resilienceEnsIm}
\end{figure}

In this scenario, the performance of the trained models was measured after randomly removing the BSSIDs, gradually and one at a time. The results are presented in Figure \ref{fig:resilienceEnsIm}, showing adequate resilience. The neural network and nearest neighbors in particular show good performance for up to 150 missing BSSIDs, whereas all models perform well for up to 50 missing BSSIDs.
Apart from the minor improvement on overall performance, model ensembling showed a significant positive impact on resilience. The ensemble's performance continued to perform best across all cases of missing number of WiFis. The ensemble's performance increases with the distance from individual model around the 250 missing WiFis mark.

Moving further to the right of this mark, we observe that GBM bounces back and performs better than the other two models. However, all scores are poorer, with an error greater than 6m (3 times more than the original). Although this error may be acceptable in some use cases (e.g. finding a vulnerable, mobility impaired user, meet and greet), it may not be acceptable in scenarios such as social distancing or identification of number of people in a particular room to send automated warnings to users to take action and increase their distance.

\subsubsection{Impact of target grid on overall accuracy}
The target grid spacing on the testbed is approximately $2.32$m. By removing every second intermediate node, the data set is split into two new data sets with different, no overlapping target nodes. As a result, each data set defines a grid with a $3.91$m spacing. The results presented in this subsection are the average of two training pipelines, one for each sub-dataset.

\begin{table}[width=.9\linewidth,cols=4,pos=h]
\caption{Performance of different models with larger target spacing}
\label{table:model_performance_double_target_space}
\begin{tabular*}{\tblwidth}{@{} LCCC@{} }
\toprule
 Model & same split & other split & average\\
 \midrule
 GBM & 1.89  & 2.17 & 2.03\\
 RNN & 1.92 & 2.46 & 2.19\\
 KNN & 2.37 & 2.56 & 2.46 \\  
 \midrule
 GBM + RNN & 1.82 & 2.18 & 2.00\\ 
 2*GBM + RNN & 1.83 & 2.14 & 1.99\\
 \bottomrule
\end{tabular*}
\end{table}

\begin{figure}[h]
    \centering
    \includegraphics[width=0.48\textwidth, trim=20 20 0 23, clip]{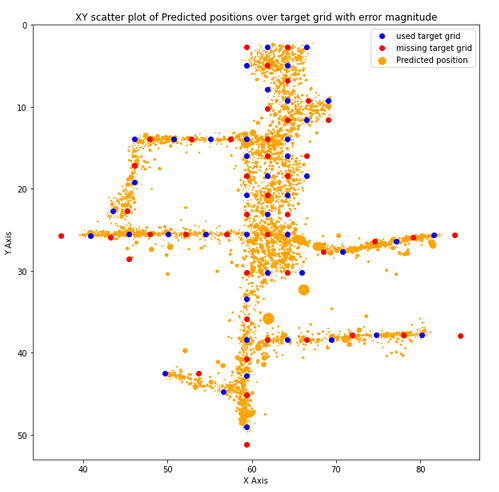}
    \caption{The effect of the two complementary grid sets on prediction (the radius of the orange dot is relative to the estimation error).}
    \label{fig:halfgrid_set2_magnIm}
\end{figure}

After completion of training with half the original data and with a more sparse target grid, the performance of each model and the ensemble model decreased as expected. As shown in Table \ref{table:model_performance_double_target_space}, the first column (same split) shows the model performance for the data belonging to the same split and the second column (other split) shows the performance for the data from the complementary split, with the training target nodes missing. The third column is the average of these two columns representing the true model performance in this setup. Figure \ref{fig:halfgrid_set2_magnIm} shows the two sets of targets (blue and red dots) as well as individual predictions (of the same split) of magnitude, where the larger the dot, the larger the prediction error. 

\section{Conclusions and future work}
\label{conclusion}
Indoor location tracking is a valuable primitive function for a number of use cases. It was demonstrated in this research that indoor location awareness while respecting sustainability principles is attainable by employing MCS and machine learning. Indoor location prediction with less than 2m error was achieved by using data from available WiFi signals only. The three models were trained out of which both GBM and RNN gave best performance for the task. Ensembling GBM with RNN not only increased the overall performance but also resulted in a significant improvement in resilience. 

While the testbed consisted of the ground floor of a relatively large building with a diverse layout, future experiments could leverage the data annotation and training protocols proposed in this work to construct and explore more complex structures and paths (such as moving along the Z axis by taking a lift or a staircase), as well as scenarios such as crowd management, real-time social distancing, dynamic fire evacuation plans, including the support of vulnerable citizens.

The dataset used for this research is made publicly available to facilitate further research and contributions. Apart from the WiFi signals, it also contains data from the internal sensors (accelerometer, gyroscope, rotation, magnetometer) that, when combined with the annotated grid points, they can be used to promote research on dead reckoning techniques, offering even higher levels of resilience to indoor location capabilities.

\section*{Acknowledgment}
This work has been partially supported by IDEAL-CITIES; a European Union’s Horizon 2020 research and innovation staff exchange programme (RISE) under the Marie Skłodowska-Curie grant agreement No 778229.


\bibliographystyle{cas-model2-names}

\bibliography{refs.bib}


\end{document}